\documentclass{article}
\usepackage[utf8]{inputenc}
\usepackage{graphicx}
\usepackage{xcolor}
\usepackage[ruled,vlined]{algorithm2e}
\usepackage{mathrsfs}
\usepackage{mathtools}
\usepackage{amsfonts}
\usepackage{amsmath}
\usepackage[a4paper, total={6in, 9in}]{geometry}
\usepackage[numbib]{tocbibind}
\DeclareMathOperator{\tr}{tr}
\DeclareMathOperator*{\argmin}{argmin}

\title{
DOODLER: Determining Out-Of-Distribution\\ Likelihood from Encoder Reconstructions
\bigskip
}

\author{
By: Jonathan S. Kent\\
\texttt{jskent2@illinois.edu}
\and
Advisor: Professor Bo Li
}
\date{\today}

\begin{document}
\vbox{
    \vspace{1in}
    \centering
    \includegraphics[width=0.3\textwidth]{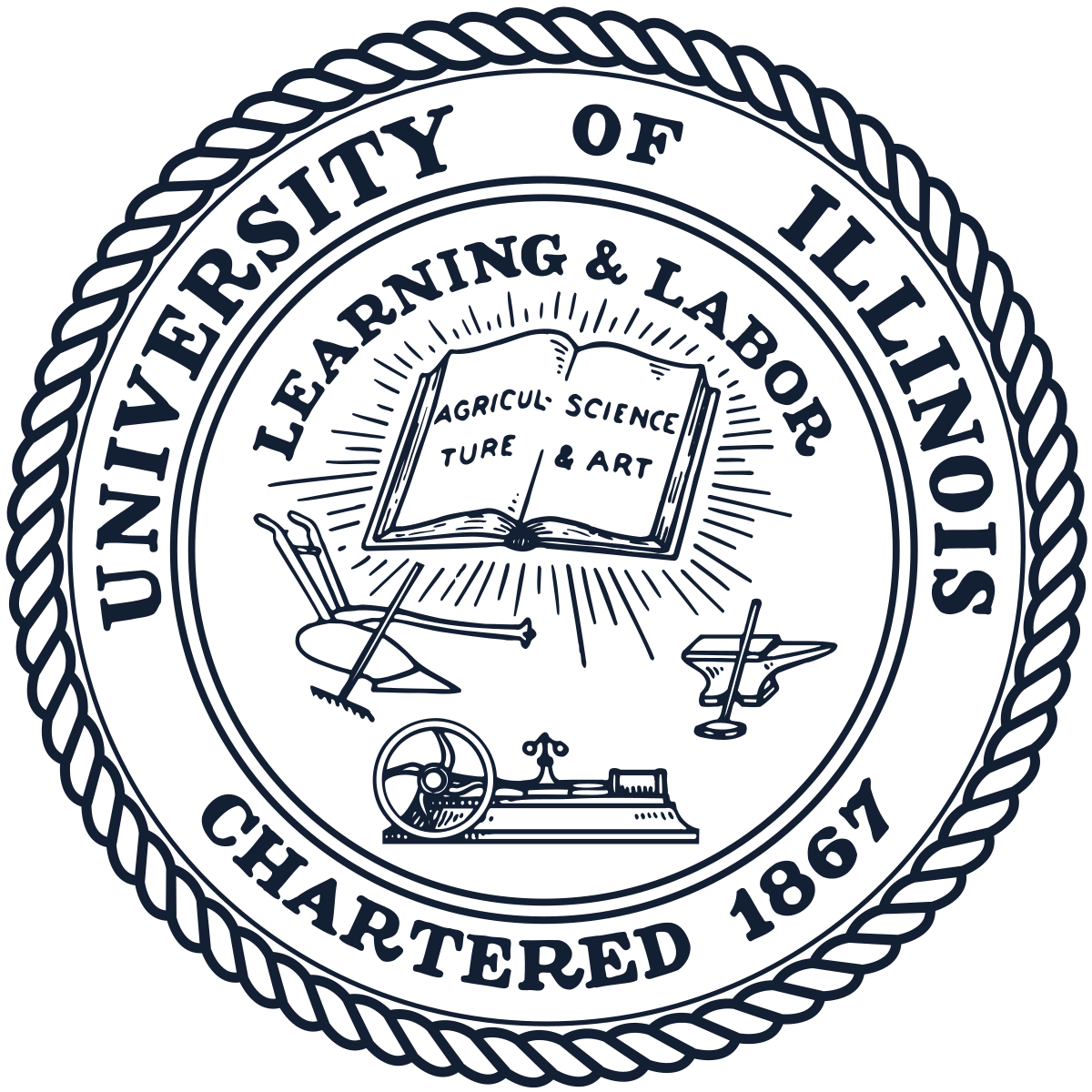}\\
    \bigskip
    \large \textbf{University of Illinois at Urbana-Champaign}\\
    \textit{Department of Mathematics}\\
    \textit{Department of Computer Science}\\
    \hrulefill\\
    \bigskip
    UNDERGRADUATE THESIS\\
    Submitted as part of an undergraduate research program\\
    \maketitle
}
\newpage

\tableofcontents

\section*{Acknowledgements}

I would like to begin by thanking my mom and dad for their unending and essential support, and all the other obvious reasons why anyone would thank their parents. But I must also thank them, Elliot N. Kent and Marsha N. Hahn, as individuals, unique and especial, because I get to be nobody else but their son. They taught me to be me. They taught me to learn. It is from them that I inherited my appetite for knowledge, and my desire to bend that knowledge to the world as it exists. What I have written here, being both original Scientific research and working towards real-world applications, is dedicated to them.\\
\\
I would also like to thank my friend, Moira E. Iten, and her family, for their ceaseless moral support and cheering-on. It would be impossible to determine the precise extent of their contributions, but that is meaningless in the face of my appreciation for them.\\
\\
I will finish by thanking my colleagues: Charles C. Wamsley, Davin Flateau, and Amber Ferguson, and all the other folks at Ball Aerospace \& Technologies Corp., for the excellent and interesting technical conversations, the insight into the problems faced in the real-world practice of Machine Learning, and the opportunity to develop as a Researcher and Engineer.

\begin{center}
\scalebox{.2}{\textcolor{white}{``A little song, a little dance, a little seltzer down your pants." -Tim}}
\end{center}
\newpage

\begin{figure}[t]
    \centering
    \includegraphics[width=0.95\textwidth]{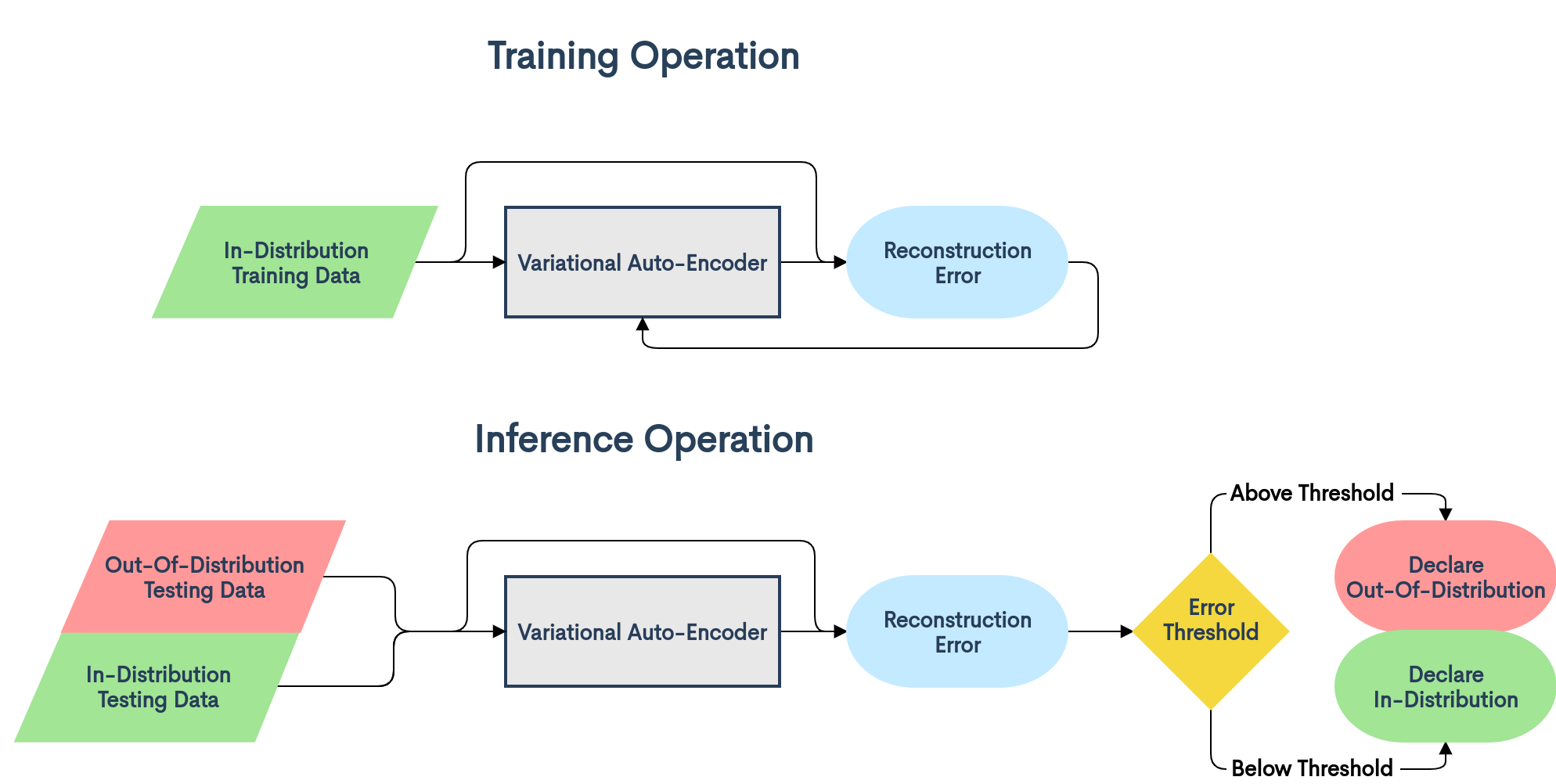}
    \caption{Training and inference steps for using DOODLER}
    \label{fig:doodlermethod}
    \hrulefill
\end{figure}

\section{Abstract}

Deep Learning models possess two key traits that, in combination, make their use in the real world a risky prospect. One, they do not typically generalize well outside of the distribution for which they were trained, and two, they tend to exhibit confident behavior regardless of whether or not they are producing meaningful outputs. While Deep Learning possesses immense power to solve realistic, high-dimensional problems, these traits in concert make it difficult to have confidence in their real-world applications. To overcome this difficulty, the task of Out-Of-Distribution (OOD) Detection has been defined, to determine when a model has received an input from outside of the distribution for which it is trained to operate.\\
\\
This paper introduces and examines a novel methodology, DOODLER, for OOD Detection, which directly leverages the traits which result in its necessity. By training a Variational Auto-Encoder (VAE) on the same data as another Deep Learning model, the VAE learns to accurately reconstruct In-Distribution (ID) inputs, but not to reconstruct OOD inputs, meaning that its failure state can be used to perform OOD Detection. Unlike other work in the area, DOODLER requires only very weak assumptions about the existence of an OOD dataset, allowing for more realistic application. DOODLER also enables pixel-wise segmentations of input images by OOD likelihood, and experimental results show that it matches or outperforms methodologies that operate under the same constraints. 

\section{Introduction}

Perhaps the most fundamental of all questions in both the academic study and applied practice of Machine Learning is that of Out-Of-Distribution Generalization, or how to produce a model that is capable of operating in settings for which it has not been expressly prepared \cite{zhou2021domain}. This is an extraordinarily difficult problem, and while inroads have been made in some areas like Online Learning and Meta-Learning \cite{sahoo2017online, zhang2020adaptive}, it is still generally true that

\begin{enumerate}
    \item Machine Learning models are not good at generalizing outside of the domain they were trained for, and
    \item Machine Learning models will still express their inferences with high confidence when outside their domain \cite{meinke2019towards}.
\end{enumerate}

\noindent The confluence of these traits means that, despite the extraordinary power of Machine and Deep Learning to crack complex, real-world, hyperdimensional problems that are intractable with any other methodology \cite{poggio2017and}, actually using these models in a production environment becomes a fraught decision. This is true to an extreme degree in areas like Vehicle Autonomy, Medical Diagnostics, and Defense, where a wrong decision made confidently may result in human casualties.\\
\\
Therefore, robust Out-Of-Distribution Detection, the ability to preempt the differences between laboratory and real-world conditions, to determine when a model is receiving data for which it is not prepared, is essential. By making these determinations accurately, it allows for autonomous systems to activate a backup protocol or Human-in-the-Loop system, greatly increasing their trustworthiness for applications.

\section{Related Work}

Much of the OOD Detection research literature available makes a number of strange assumptions; among these, the existence of a dataset consisting of Out-Of-Distribution samples to train a binary classifier \cite{meinke2021provably}. To comprehensively cover what is OOD, this dataset would have to contain samples from the entirety of the space outside of the In-Distribution. If the ID contained pictures of dogs, for example, this OOD dataset would have to contain pictures of literally everything that isn't a dog: cats, foxes, centaurs, Gaussian noise, fruit, supernovae, cartographic maps of Poland, so on and so forth. If any class of data is left out, then it is outside of the distribution for which the detector is prepared, making it necessary to perform OOD Detection in order to perform OOD Detection, leaving us back at square one. Thus, any true OOD Detector must be trained in an Unsupervised manner, without training on an OOD dataset.\\
\\
Other methods, like ODIN \cite{liang2020enhancing}, make significant progress towards achieving this. ODIN operates by comparing a classification model's output on an input to its output when the input has a small adversarial perturbation, with the output changing less for Out-Of-Distribution inputs than In-Distribution ones. While this performs very well, it's also much more computationally expensive, requiring two forward passes and a gradient calculation, than single-shot inference, so it's unsuited to real-time applications. Also worth noting is that, although the model may be trained without OOD samples, the analysis of the output distributions required to differentiate between ID and OOD data requires at least a few samples. But this is a much weaker requirement than a comprehensive OOD dataset, as it's only needed in order to determine a handful of parameters, rather than to train a model that takes them as input. As a result, ODIN can be assumed to generalize much more reasonably.\\
\\
Another method, DAGMM \cite{zong2018deep}, involves using a Gaussian Mixture Model on the latent space of a Variational Auto-Encoder \cite{kingma2014autoencoding}. While a classical model like a Gaussian Mixture would be unable to deal with the high dimensionality of images, the low number of dimensions of the VAE's encodings means that it can handle much more abstract data than might otherwise be possible. This still makes a number of assumptions about the underlying distribution, including that the probability density of the latent space can be represented meaningfully by a Gaussian Mixture Model. It also imposes both a limit on the number of dimensions that can be used in the latent space, and a requirement that this number of dimensions stays constant between training, testing, and real-world inference. This is disappointing, as ordinarily fully convolutional models can be trained on smaller chips than they will be used for, making training easier. However, using VAEs for OOD Detection does get at the heart of the matter, because the underlying problem involves getting a handle on patterns in the In-Distribution, which is fundamental to how VAEs operate.
\begin{table}[!t]
    \centering
    \begin{tabular}{c|c|c|c}
                              & ODIN         & DAGMM        & DOODLER      \\
        \hline
        Unsupervised          & $\checkmark$ & $\checkmark$ & $\checkmark$ \\
        Chip Size Invariant   & $\checkmark$ &              & $\checkmark$ \\
        One-Shot Inference    &              & $\checkmark$ & $\checkmark$ \\
        Segmentation          &              &              & $\checkmark$ \\
        \hline
    \end{tabular}
    \caption{Feature Comparison between ODIN, DAGMM, and DOODLER}
    \label{tab:features}
    \hrulefill
\end{table}
\\
\\
Something else worth considering is the possibility of performing segmented Out-Of-Distribution Detection on images, to help with explainability; an OOD Detector would be much more trustworthy, and provide helpful information to a Human-in-the-Loop counterpart, if it could point out exactly where in an image it was seeing something that set it off. However, the work in this area has relied extensively on the existence of OOD datasets \cite{bevandic2018discriminative, hendrycks2019benchmark, williams2021fool}, so it could still benefit greatly from moving to an Unsupervised setting.\\
\\
But on a fundamental level, much of the available Unsupervised Out-Of-Distribution Detection research comes at it from an analytic rather than a functional perspective. Broadly, the work on the topic has thought of the defining trait of OOD data to be that they literally fall outside of the high-density region of a probabilistic In-Distribution. However, when looking at it from a functional perspective, the primary trait of OOD data is that they cause models to fail and fail confidently, this being the reason why it is valuable to detect them in the first place. By shifting over to this framework, a new methodology immediately presents itself: find some task for which a failure state can be detected without a ground truth label for comparison, and then use that failure state as a detection signal. Thus, DOODLER: Determining Out-Of-Distribution Likelihood from Encoder Reconstructions.

\section{Methodology}

The DOODLER methodology we propose, illustrated in Figure \ref{fig:doodlermethod}, involves training a Variational Auto-Encoder on the same training data as the Deep Learning model you want to perform Out-Of-Distribution Detection for, and then setting conditions on the reconstruction error under which you declare inputs to be OOD. Compared to its closest relatives, ODIN and DAGMM, DOODLER is similarly unsupervised, but improves on DAGMM by being invariant to chip size, improves on ODIN by only requiring one forward pass per input, and improves on them both by enabling image segmentation, illustrating which regions specifically are OOD within an image.

\subsection{Variational Auto-Encoder}

Per \cite{kingma2014autoencoding}, let's assume that a given sample $x_i$ is generated from a process involving a lower-dimensional latent representation $z_i$. Where $x_i$ belongs to an high-dimensional field $\mathcal{X}$, $z_i$ belongs to a low-dimensional field $\mathcal{Z}$, such that $\dim(\mathcal{X}) >> \dim(\mathcal{Z})$. If $x_i$ is, say, a three-channel RGB $H \times W$ picture of a dog, then $x_i \in \mathcal{X} = \mathbb{R}^{3 \times H \times W}$, while $z_i \in \mathcal{Z} = \mathbb{R}^m$ is a vector in ``Dog-Space," with $m$ dimensions describing higher-order concepts like fur length, appearance, posture, and breed.\\
\\
This allows us to define the process $f : \mathbb{Z} \to \mathbb{X}$ by which $z_i$ becomes $x_i$, by saying that $f(z_i) \coloneqq x_i$. In the context of dog photos, $f$ would involve taking the vector description $z_i$ of a dog, and using it to produce a picture of the dog described. Furthermore, we might hypothesize that $f$ is invertible, with inverse $g$, such that $f\big(g(x_i)\big) = x_i$. Here, $g$ would start by taking a picture of a dog, and then producing the vector description of that dog, with the condition that this vector can be used by $f$ to reconstruct the original image of the dog.\\
\\
However, while $\mathcal{X}$, containing photographs, can be reasonably understood by humans, the field $\mathcal{Z}$, containing the perfect description of dogs, and the process $f$, by which a dog is brought into existence and photographed, are almost certainly inscrutable. What higher-order concepts might a human overlook when discussing a dog or its appearance? How could they precisely describe the process by which a dog might be drawn? So, $\mathcal{Z}$ and $f$ will instead be approximated via a learning algorithm, using $\hat{\mathcal{Z}} = \mathbb{R}^m$ with some manually selected $m$ as a stand-in for $\mathcal{Z}$, and a neural network $\hat{f}$ parameterized with $\theta_{\hat{f}}$ of the form $\hat{f}(\theta_{\hat{f}}; \hat{z}_i) = \hat{x}_i$. This will allow for the model $\hat{f}(\theta_{\hat{f}}; \cdot)$ to be created automatically by training it on data, and together with a second neural network of the form $\hat{g}(\theta_{\hat{g}}; x_i) = \hat{z}_i$ to invent a meaningful version of $\hat{\mathcal{Z}}$. In a sense, by training these two neural networks on photographs of dogs, the requirement that $\dim(\mathcal{X}) >> \dim(\hat{\mathcal{Z}})$ means that $\hat{g}(\theta_{\hat{g}}; \cdot)$ and $\hat{f}(\theta_{\hat{f}}; \cdot)$ will have to learn to produce a meaningful description of a dog's fur or posture in a higher-order, lower-dimensional setting than a picture, and to then re-draw that picture based on that description, respectively.\\
\\
By composing $\hat{f}(\theta_{\hat{f}}; \cdot)$ and $\hat{g}(\theta_{\hat{g}}; \cdot)$, they can be referred to as a single model, the Variational Auto-Encoder $h(\theta; \cdot) \coloneqq \hat{f}\big(\theta_{\hat{f}}; \hat{g}(\theta_{\hat{g}}; \cdot)\big)$, with $\theta = \theta_{\hat{f}} \cup \theta_{\hat{g}}$. In order to train this combined model $h(\theta; \cdot)$, we will use a training dataset $\mathcal{D}_{IDtrain} \coloneqq \{x_i\} \sim \mathscr{D}_{ID}$, which consists of samples from a distribution over the input field $\mathcal{X}$, and then attempt to minimize $\mathbb{E}\Big[\mathcal{L}\big(h(\theta; x_i), x_i\big)\Big]$ for some loss or reconstruction error function $\mathcal{L}(\hat{x}_i, x_i)$. To use DOODLER to detect Out-Of-Distribution inputs for an application, $\mathcal{D}_{IDtrain}$ would be the same dataset that the application model was trained on - here, it would contain pictures of dogs - and $\mathcal{L}$ might be something like pixel-wise Mean Squared Error, which we will justify using in the next section. To optimize the parameters of the VAE, an algorithm like Adam Optimizer might be used, denoted as $\mathcal{A}lg$. This training procedure is given by Algorithm \ref{alg:trainvae}.\\
\begin{algorithm}[t]
\SetAlgoLined
\KwResult{A Variational Auto-Encoder $h(\theta; \cdot)$}
Initialize $\theta_0$ randomly, $t = 1$\;
\While{$||\theta_t - \theta_{t-1}|| > \epsilon$}{
    $X_t \leftarrow \{x_0, x_1, \dots, x_n\} \sim \mathcal{D}_{IDtrain}$\;
    $\hat{X}_t \leftarrow \{\hat{x}_0, \hat{x}_1, \dots, \hat{x}_n\} = \{h(\theta_t; x_i) | x_i \in X_t\}$\;
    $G_t = \nabla_\theta \frac{1}{n} \sum_{i = 0}^n \mathcal{L}(\hat{x}_i, x_i)$\;
    $\theta_{t + 1} \leftarrow \mathcal{A}lg(\theta_t, G_t)$\;
    $t \leftarrow t + 1$\;
}
\caption{Variational Auto-Encoder Training Procedure}
\label{alg:trainvae}
\end{algorithm}
\\
This produces a model $h(\theta; \cdot)$ that, in some sense, has an understanding of the structure of $\mathscr{D}_{ID}$. If this understanding was not present, then $\hat{g}(\theta_{\hat{g}}; \cdot)$ would not be able to encode an input from $\mathscr{D}_{ID}$ into a lower dimension and then have $\hat{f}(\theta_{\hat{f}}; \cdot)$ successfully reconstruct it, so they must be leveraging context and patterns within the data to compress and recover information. Because the lower dimension of the compressed state means that some information will be lost, the information that is preserved must be capable of being understood in such a way as to fill in most of the missing information. If trained on pictures of dogs, by understanding what dogs look like $\hat{g}(\theta_{\hat{g}}; \cdot)$ might encode information like ``the tail is at this location, and is upright with long golden-brown fur" in order for $\hat{f}(\theta_{\hat{f}}; \cdot)$ to reconstruct the pixel information of the region containing the tail.\\
\\
However, since this model operates entirely on the basis of understanding the structure of $\mathscr{D}_{ID}$, if it was provided with $x_i$ sampled from somewhere else, possessing structures and patterns that differ greatly from $\mathscr{D}_{ID}$, then it's reasonable to assume that $\mathcal{L}\big(h(\theta; x_i), x_i\big)$ would be relatively large. For example, if a photograph of an airplane was put into the Variational Auto-Encoder trained on pictures of dogs, we could guess that $\hat{g}(\theta_{\hat{g}}; \cdot)$ would not be able to successfully encode the appearance of a jet engine into dimensions like ``fur length" and ``ear size," instead producing numerical gibberish. From there, $\hat{f}(\theta_{\hat{f}}; \cdot)$ would attempt to turn that gibberish into a picture of some kind of weird dog, resulting in significant error when compared to the original airplane.

\subsection{Reconstruction Error}

Within the framework of \cite{kingma2014autoencoding}, a ``canonical" Variational Auto-Encoder consists of two parts, $p(\theta_p; \cdot | \cdot)$ and $q(\theta_q; \cdot | \cdot)$, which are conditional probability estimators, with $q(\theta_q; \hat{z}_i | x_i)$ being the estimated probability that $\hat{z}_i$ is the true encoding of $x_i$, and $p(\theta_p; x_i | \hat{z}_i)$ being the estimated probability of receiving $x_i$ given the encoding $\hat{z}_i$. This means that training $p(\theta_p; \cdot | \cdot)$ and $q(\theta_q; \cdot | \cdot)$ involves reducing the degree to which they disagree for any given sample $x_i$. For example, if $q(\theta_q; \hat{z}_i | x_i)$ is high, meaning $q(\theta_q; \cdot | \cdot)$ estimates that $\hat{z}_i$ is likely the true encoding of $x_i$, then $p(\theta_p; x_i | \cdot)$ should be low everywhere besides $\hat{z}_i$. This disagreement over $\hat{z}_i$ when considering $x_i$ is expressed as the Kullback-Leibler Divergence:

$$
D_{KL}(\theta; x_i) = \mathbb{E}_{q(\theta_q; \hat{z}_i|x_i)}\big[\ln p(\theta_p;x_i|\hat{z}_i) - \ln q(\theta_q; \hat{z}_i|x_i)\big]
$$

\noindent However, actually estimating $D_{KL}(x_i)$ and its gradients is computationally infeasible. Going forward, we'll be abbreviating the notation using $p = p(\theta_p;x_i|\hat{z}_i)$ and $q = q(\theta_q; \hat{z}_i|x_i)$, to make further algebraic work more legible. By using the same assumption as \cite{kingma2014autoencoding} that the distributions over $\hat{z}_i$ represented by $p$ and $q$ approximate multivariate Gaussian distributions, we get the closed form:\footnote{Here, as in standard Statistical notation, $\Sigma$ is the covariance matrix and $\mu$ is the vector-valued mean of a multivariate distribution. Additionally, `$\tr$' and `$\det$' are the trace and determinant of a matrix, and `$\dim$' is the dimension of a field.}

$$D_{KL}(\theta; x_i) = \frac{1}{2}\Bigg(
\tr\big(\Sigma_p^{-1}\Sigma_q\big) + 
(\mu_p - \mu_q)^T\Sigma_p^{-1}(\mu_p - \mu_q) - 
\dim(\mathcal{\hat{Z}}) + 
\ln\Big(\frac{\det \Sigma_p}{\det \Sigma_q}\Big)
\Bigg)
$$

\noindent Next, it's worth noting that a nonzero covariance between different dimensions in $\mathcal{\hat{Z}}$ in either distribution would mean that the same information was being stored in multiple entries in $\hat{z}_i$. Because the number of entries is very limited, this repetition in storage is expected to disappear, and the covariance matrices will become approximately diagonal. By replacing the covariance matrices $\Sigma$ with just their diagonal entries in the form of variance vectors $\sigma^2$, and by removing the constant due to the dimension of $\mathcal{\hat{Z}}$, we can approximate and simplify this expression without seriously affecting the future gradients with respect to $\theta$:

$$
\widetilde{D}_{KL}(\theta; x_i) = \frac{1}{2}\Big(
\sigma_p^{-2} \cdot \sigma_q^2 + 
\sigma_p^{-2} \cdot (\mu_p - \mu_q)^2 +
||\ln\sigma_p^2||_1 - ||\ln\sigma_q^2||_1
\Big)
$$

\noindent Lastly, we can insert a normalization step into the modeling process itself, forcibly setting the variances along each dimension to 1. This guarantee reduces the entire expression to:

$$
\widetilde{D}_{KL}(\theta; x_i) = \frac{1}{2}||\mu_p - \mu_q||_2^2
$$

\noindent What this means is that, effectively, when given $x_i$, the gradient behavior of the Kullback-Leibler Divergence between $p(\theta_p;x_i|\cdot)$ and $q(\theta_q; \cdot|x_i)$ is similar to that of the $L2$ distance between their means in $\mathcal{\hat{Z}}$. By moving back into thinking in terms of functions, with $\hat{f}(\theta_{\hat{f}}; \hat{z}_i) = \hat{x}_i$ and $\hat{g}(\theta_{\hat{g}}; x_i) = \hat{z}_i$, we can use the local continuity of $\hat{f}(\theta_{\hat{f}}; \cdot)$ and $\hat{g}(\theta_{\hat{g}}; \cdot)$ to get:

$$
||\hat{f}^{-1}(\theta_{\hat{f}}; x_i) - \hat{g}(\theta_{\hat{g}}; x_i)||_2^2 \propto ||\hat{f}(\theta_{\hat{f}}; \hat{g}(\theta_{\hat{g}}; x_i)) - x_i||_2^2
$$

\noindent Meaning that it's possible to relate the Kullback-Leibler Divergence over $\mathcal{\hat{Z}}$ to the Mean Squared Error in $\mathcal{X}$, reducing to a loss function given by:

\begin{equation}
    \mathcal{L}(\hat{x}_i, x_i) = \frac{||\hat{x}_i - x_i||^2_2}{\dim(\mathcal{X})}
    \label{eqn:loss}
\end{equation}

\noindent Which satisfies the condition that the value of $\hat{x}_i$ which minimizes $\mathcal{L}(\hat{x}_i, x_i)$ is the same as the mean of $p(\theta_p;\cdot)$ given $q(\theta_q;\hat{z}_i|x_i)$ with optimal $\theta$, or

$$
\argmin_{\hat{x}_i \in \mathcal{X}}\mathcal{L}(\hat{x}_i, x_i) = 
\mu \bigg(
p\Big(\theta_p;x_i|q(\theta_q; \hat{z}_i|x_i)\Big) \bigg|
\theta = \argmin_{\theta \in \Theta}D_{KL}(\theta, x_i)
\bigg)
$$

\noindent Therefore, the mean of the squared differences between $x_i$ and $\hat{x}_i = h(\theta;x_i)$ has gradient behavior that is approximately equivalent to that of the Kullback–Leibler formulation, and can be used both as a loss function for the direct optimization of the Variational Auto-Encoder for Algorithm \ref{alg:trainvae}, and as the reconstruction error during inference.


\section{Error Distribution Analyses}

\begin{figure}[t]
    \centering
    \includegraphics[width=0.6\textwidth]{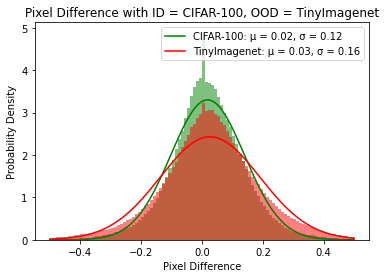}
    \caption{Distribution of Pixel-Wise Differences between example $x_i$ and $\hat{x}_i$}
    \label{fig:pixelwise}
    \hrulefill
\end{figure}

In order to find the best rules to use to declare samples to be Out-Of-Distribution, we have to know the probability distributions of the reconstruction errors. As we see in Figure \ref{fig:pixelwise}\footnote{This histogram was produced with 80 samples each from the testing sets of CIFAR-100 and TinyImagenet, using a VAE trained on CIFAR-100. This behavior, with the pixel-wise differences approximating Gaussian distributions, is general between different datasets.}, the pixel-wise differences between $x_i$ and $\hat{x}_i$ approximate a Gaussian distribution. As a result, the reconstruction error of an image, given by $l = \mathcal{L}(\hat{x}_i, x_i) = \frac{1}{\dim(\mathcal{X})}||\hat{x}_i - x_i||^2_2$ from Equation \ref{eqn:loss}, being the mean of the squares of values sampled from what is approximately Gaussian, approximates a Gamma distribution.\footnote{Formally, a Gamma distribution represents the probability of receiving a given sum when adding the squares of a known number of values sampled from a Gaussian distribution. But, because a mean is the sum over a set divided by its size, and a Gaussian divided by a constant is another Gaussian, the mean of the squares follows a Gamma distribution, the same as their sum.} This aligns well with the Gamma distribution-esque appearances of the histograms of image-wise reconstruction errors, as in Figure \ref{fig:cifar100vstinyimagenetpixels}.\\
\\
Taking the $\alpha, \beta$ formulation of the Gamma distribution, we can give the probability density function as:

$$p(l) = \frac{\beta^\alpha}{\Gamma(\alpha)}l^{\alpha-1}e^{-\beta l}$$

\noindent With $\Gamma$ being the Gamma function. Because the mean and the variance of the Gamma distribution are known to be $\frac{\alpha}{\beta}$ and $\frac{\alpha}{\beta^2}$ respectively, and because the mean and variance of the distribution of $l$ can be approximated as $\mu_l = \mathbb{E}[l]$ and $\sigma_l^2 = \mathbb{E}[l^2] - \mathbb{E}[l]^2$ to a high degree of precision with a large enough number of samples, $\alpha$ and $\beta$ can be solved for using the Method of Moments:

$$
\mu_l = \frac{\alpha}{\beta},\ \sigma_l^2 = \frac{\alpha}{\beta^2} \rightarrow
\alpha = \frac{\mu_l^2}{\sigma_l^2},\ \beta = \frac{\mu_l}{\sigma_l^2}
$$

\noindent Given a VAE $h(\theta;\cdot)$ trained on the training set $\mathcal{D}_{IDtrain}$ sampled from $\mathscr{D}_{ID}$, it is now possible to produce a distribution $p(l|x_i \sim \mathscr{D}_{ID})$, the probability density of receiving $l$ as the reconstruction error if we know that $x_i$ was sampled from the In-Distribution. We do this by taking a test set $\mathcal{D}_{IDtest}$ that has also been sampled from $\mathscr{D}_{ID}$, calculating the mean and variance of $l$ over $\mathcal{D}_{IDtest}$, and then using them to solve for $\alpha_{ID}$ and $\beta_{ID}$. Using the same method, but a test set $\mathcal{D}_{OODtest}$ that has been sampled from $\mathscr{D}_{OOD}$, we can solve for $\alpha_{OOD}$ and $\beta_{OOD}$, which parameterize $p(l|x_i \sim \mathscr{D}_{OOD})$, the probability density of $l$ when $x_i$ was sampled from Out-Of-Distribution.

\subsection{Out-Of-Distribution Sample Detection}

\begin{table}[t]
    \centering
    \begin{tabular}{c|c}
        True Positive  & $l > T_v$ and $x_i \not \sim \mathscr{D}_{ID}$\\
        False Positive & $l > T_v$ and $x_i \sim \mathscr{D}_{ID}$\\
        False Negative & $l < T_v$ and $x_i \not \sim \mathscr{D}_{ID}$\\
        True Negative  & $l < T_v$ and $x_i \sim \mathscr{D}_{ID}$
    \end{tabular}
    \caption{Conditions under which a sample $x_i$ with reconstruction error $l$ is put in each category.}
    \hrulefill
    \label{tab:defprop}
\end{table}

\begin{table}[t]
    \centering
    \begin{tabular}{ccc}
        $
\mathcal{TP}(T_v)$ & $=$ & $p(x_i \not \sim \mathscr{D}_{ID})\int_{l > T_v}p(l|x_i \not \sim \mathscr{D}_{ID}) \partial l
$\\
        $
\mathcal{FP}(T_v)$ & $=$ & $p(x_i \sim \mathscr{D}_{ID})\int_{l > T_v}p(l|x_i \sim \mathscr{D}_{ID}) \partial l
$\\
        $
\mathcal{FN}(T_v)$ & $=$ & $p(x_i \not \sim \mathscr{D}_{ID})\int_{l < T_v}p(l|x_i \not \sim \mathscr{D}_{ID}) \partial l
$\\
        $
\mathcal{TN}(T_v)$ & $=$ & $p(x_i \sim \mathscr{D}_{ID})\int_{l < T_v}p(l|x_i \sim \mathscr{D}_{ID}) \partial l
$
    \end{tabular}
    \caption{Calculating proportions of the data}
    \hrulefill
    \label{tab:calcprop}
\end{table}

\begin{table}[t]
    \centering
    \begin{tabular}{ccc}
    $Sensitivity(T_v)$   & $=$ & $\frac{\mathcal{TP}(T_v)}{\mathcal{TP}(T_v) + \mathcal{FN}(T_v)}$\\
    $Specificity(T_v)$   & $=$ & $\frac{\mathcal{TN}(T_v)}{\mathcal{TN}(T_v) + \mathcal{FP}(T_v)}$\\
    $\mathcal{PPV}(T_v)$ & $=$ & $\frac{\mathcal{TP}(T_v)}{\mathcal{TP}(T_v) + \mathcal{FP}(T_v)}$\\
    $\mathcal{NPV}(T_v)$ & $=$ & $\frac{\mathcal{TN}(T_v)}{\mathcal{TN}(T_v) + \mathcal{FN}(T_v)}$\\
    \end{tabular}
    \caption{The values derived from the proportions of the data.}
    \hrulefill
    \label{tab:derprop}
\end{table}

If some prior probabilities $p(x_i \sim \mathscr{D}_{ID}) + p(x_i \not \sim \mathscr{D}_{ID}) = 1$ are assumed, as well as the existence of a relatively small Out-Of-Distribution dataset $\mathcal{D}_{OODtest}$ sampled from $\mathscr{D}_{OOD}$ to determine $p(l|x_i \sim \mathscr{D}_{OOD})$, it is possible to use Bayes' Theorem to compute $p(x_i \sim \mathscr{D}_{ID}|l)$ with

$$
p(x_i \sim \mathscr{D}_{ID}|l) = \frac{p(l|x_i \sim \mathscr{D}_{ID}) \cdot p(x_i \sim \mathscr{D}_{ID})}{p(l|x_i \sim \mathscr{D}_{ID}) \cdot p(x_i \sim \mathscr{D}_{ID}) + p(l|x_i \sim \mathscr{D}_{OOD}) \cdot p(x_i \not \sim \mathscr{D}_{ID})}
$$

\noindent At which point $p(x_i \sim \mathscr{D}_{ID}|l)$ going below some threshold probability $T_p$ would be used to declare a positive Out-Of-Distribution Detection. However, it must be noted that this does rely on both the existence of $\mathcal{D}_{OODtest}$, and on the approximate equivalence between $p(l|x_i \sim \mathscr{D}_{OOD})$ and $p(l|x_i \not \sim \mathscr{D}_{ID})$. But, these are much weaker assumptions than those made by works like \cite{meinke2019towards}, as we're not using a dataset sampled from $\mathscr{D}_{OOD}$ for training $h(\theta; \cdot)$, reducing the quantity of data needed. We only ever use $\mathcal{D}_{OODtest}$ to approximate the parameters of Statistical distributions, not to capture the actual underlying behavior of $x_i \not \sim \mathscr{D}_{ID}$.\\
\\
For a given threshold probability $T_p$, it's possible for us to calculate the expected True Positive, False Positive, False Negative, and True Negative proportions of the inferences made. With ``positive" meaning the presence of an Out-Of-Distribution sample, these are the proportions of incoming data that were sampled Out-Of-Distribution and correctly detected, that were sampled In-Distribution and incorrectly detected as Out-Of-Distribution, that were sampled Out-Of-Distribution and incorrectly ignored as In-Distribution, and that were sampled In-Distribution and were correctly ignored, respectively. \\
\\
Under the assumption that $p(l|x_i \sim \mathscr{D}_{OOD})$, the distribution of reconstruction errors from the Out-Of-Distribution, is shifted to the right of $p(l|x_i \sim \mathscr{D}_{ID})$, the distribution from the In-Distribution, this begins by finding the matching threshold value $T_v$, using:

$$T_v = l|[p(x_i \sim \mathscr{D}_{ID}|l) = T_p]$$

\noindent And then using the conditions from Table \ref{tab:defprop} to define when a sample will fall into each category. These conditions lead to the formulae listed in Table \ref{tab:calcprop}, which can be used to calculate the proportions of the data that will fall into each category. Because those formulae, when applied to this situation, involve integrating the probability density function of the Gamma distribution, it may be useful to replace the integral with the cumulative density function:

$$
\int_a^b \frac{\beta^\alpha}{\Gamma(\alpha)}l^{\alpha-1}e^{-\beta l}\partial l = 
\frac{1}{\Gamma(\alpha)}\gamma(\alpha,\beta l) \Bigg|_{l=a}^{l=b}
$$

\noindent But this comes with the caveat that $\gamma$ is itself defined as the integral that satisfies this condition, rather than it being possible to reduce the integral to elementary functions. Next, having determined those proportions, we can calculate the Sensitivity, Specificity, Positive Predictive Value, and Negative Predictive Value, using the relationships from Table \ref{tab:derprop}. Those values in turn can be used to produce the Receiver Operating Characteristic ($ROC$) curve. This curve is generated by pairs of the False Positive Rate, or ``Fall-Out," and the True Positive Rate, or ``Recall," taken at the same threshold value. These are not the False Positive Proportion and True Positive Proportion from Tables \ref{tab:defprop} and \ref{tab:calcprop}, but rather the rates at which negative cases are declared positive and positive cases are declared positive, respectively. The False Positive Rate/Fall-Out, here $f$, is the probabilistic inverse of Specificity, and the True Positive Rate/Recall, here $r$, is equivalent to Specificity. These are used to produce the $ROC$ curve:

$$
ROC = \{(f, r) = \big(1 - Specificity(T_v), Sensitivity(T_v)\big)| T_v \in \mathbb{R}\}
$$

\noindent Which may be turned into a functional form:

$$
ROC(f) = r | (f, r) \in ROC
$$

\noindent And then integrated, to retrieve the Area Under the ROC Curve (AUROC):

$$
AUROC = \int_0^1 ROC(f)\partial f
$$

\noindent This integral itself being possible to approximate computationally by using an interpolation method on a finite subset of $ROC$ to yield $\widetilde{ROC}(f)$. Getting this number, the AUROC, is useful, as it represents the model's trade-off between the True Positive Rate and the False Positive Rate at different thresholds. A high AUROC means that the model can have both a high True Positive Rate and a low False Positive Rate at the same time, indicating that the model performs well. A similar method can be used to calculate the Area Under the Precision-Recall Curve, which measures how the model trades off between minimizing false positives, and minizing false negatives.

\subsection{Out-Of-Distribution Pixel Detection}

One of the most important usecases of Out-Of-Distribution Detection involves alerting a human operator that an OOD input has been detected, so that the human can make a more sensible decision than a model would be able to. What might be helpful for a human analyst is to understand what, specifically, about the input caused it to declared OOD. Helpfully, DOODLER can be used to produce a segmentation over an image, based on how likely it is that each individual pixel belongs to an OOD input.\\
\\
Determining the probability of a given pixel having been sampled OOD based on the reconstruction error falls along similar lines to doing for an entire sample. Collect pixel-wise reconstruction errors from both In-Distribution and Out-Of-Distribution samples, determine the parameters of the $\chi^2$ distributions that produce them, and then use Bayes' Theorem again. For a given pixel reconstruction error $t$ produced by a pixel $v_i$: 

$$
p(v_i \sim \mathscr{D}_{ID}|t) = \frac{p(t|v_i \sim \mathscr{D}_{ID}) \cdot p(v_i \sim \mathscr{D}_{ID})}{p(t|v_i \sim \mathscr{D}_{ID}) \cdot p(v_i \sim \mathscr{D}_{ID}) + p(t|v_i \sim \mathscr{D}_{OOD}) \cdot p(v_i \not \sim \mathscr{D}_{ID})}
$$

\noindent We can use $\chi^2$ distributions to approximate $p(t|v_i \sim \mathscr{D}_{ID})$ and $p(t|v_i \sim \mathscr{D}_{OOD})$, because they're the distributions of the squares of values from a single Gaussian. The $\chi^2$ distribution only has a single parameter, $\alpha$, which is the same as its mean, and a probability density function given by:

$$
p(t) = \frac{1}{2^{\alpha / 2}\Gamma(\alpha/2)}t^{\alpha/2-1}e^{-t/2}
$$

\subsection{Out-Of-Distribution Stream Detection}

\begin{table}[t]
    \centering
    \begin{tabular}{cc}
        $H_0:$ & $\mathbb{E}[l|x_i \sim S] \leq \mathbb{E}[l|x_i \sim \mathscr{D}_{ID}]$  \\
        $H_a:$ & $\mathbb{E}[l|x_i \sim S] > \mathbb{E}[l|x_i \sim \mathscr{D}_{ID}]$
    \end{tabular}
    \caption{Hypotheses for testing if a given stream is sampling Out-Of-Distribution.}
    \label{tab:hyp}
    \hrulefill
\end{table}

Compared to sample detection, stream detection doesn't even require a dataset $\mathcal{D}_{OODtest}$ or overly restrictive assumptions about $x_i \not \sim \mathscr{D}_{ID}$, as we can do it entirely through the use of Hypothesis Testing. \\
\\
For this setting, let's say that there's a stream of data $S$ from which $x_i$ is being sampled, like a sensor in a new location. Then the reconstruction error $l$ for a given sample is the result of comparing the output of the trained model $\hat{x}_i = h(\theta;x_i)$ to the sample, $l = \mathcal{L}(\hat{x}_i, x_i)$. By assuming that the expected error for Out-Of-Distribution samples is greater than it is for In-Distribution samples, the Null Hypothesis, $H_0$, and the Alternative Hypothesis, $H_a$, can be taken according to Table \ref{tab:hyp}. Essentially, we start by assuming $H_0$, that $S$ is producing ID data, with low reconstruction errors. Under this assumption, we can calculate the probability that we'd see the mean reconstruction error that $S$ has given us so far. If this probability is too low, we can reject $H_0$, saying that it's an unreasonable assumption in light of the evidence, and accept $H_a$, declaring that $S$ is sampling from Out-Of-Distribution, because it's producing higher reconstruction errors.\\
\\
The Central Limit Theorem gives us that, if we repeatedly take a number of samples from a population and calculate their mean, the distribution of the means over the repetitions approaches a normal distribution, allowing us to use Gosset's/Student's $t$-test.\footnote{This Statistical test, involving using a test statistic $t$ based on the number of standard errors that a sample mean differs from its expectated value if it was produced by a known population, was originally developed by Brewer and Statistician William Sealy Gosset, while working at Guinness. At the time, this new test allowed brewers to draw meaningful conclusions about new strains of barley or Chemical processes from smaller sample sizes. Because use of this new test for brewing constituted a trade secret, and Guinness required their researchers to write under pseudonyms, Gosset published his work under the name ``Student," after which his $t$-test and $t$-distribution are named} To start, take a set of $n$ samples:

$$\psi \coloneqq \{x_1, x_2,\dots, x_n\} \sim S$$

\noindent And take the difference between the mean reconstruction error from this set $\psi$, and the expected reconstruction error $\mu_l$ from the In-Distribution $\mathscr{D}_{ID}$:

$$
Z \coloneqq \Bigg(\frac{1}{n} \sum_{x_i \in \psi} \mathcal{L}\big(h(\theta; x_i), x_i\big)  \Bigg)
- \mu_l
$$

\noindent From there, take the standard error of the mean over $n$ samples, using the standard deviation $\sqrt{\sigma_l^2} = \sigma_l$ of the In-Distribution $\mathscr{D}_{ID}$:

$$
s \coloneqq \frac{\sigma_l}{\sqrt{n}}
$$

\noindent Which lets us calculate the test statistic:

$$
t = \frac{Z}{s}
$$

\noindent And then we can choose to reject the Null Hypothesis depending on whether or not $t$ exceeds the $Z$-score, the number of standard errors required to produce this result if the expected mean of the samples was equal to the mean of the population, corresponding to the significance level, or $P$-value. \\
\\
Under the hypothesis that $S$ really was sampling from $\mathscr{D}_{ID}$, we would expect the mean reconstruction error yielded from $n$ samples from $S$ would follow a normal distribution centered around the expected reconstruction error $\mu_l$ from $\mathscr{D}_{ID}$. This test statistic $t$ then tells us how many standard deviations above the expected reconstruction error the mean reconstruction error of your samples is. By using that number of standard deviations, together with the cumulative density of the normal distribution, we can determine the likelihood of receiving a value for $t$ as or more extreme than the one that appeared. If this likelihood is too low, say <1\%, we reject the Null Hypothesis, and say that something must be increasing the reconstruction errors of the data from $S$, namely, that they're OOD.


\section{Experimentation}

\begin{table}[t]
    \centering
    \makebox[\textwidth][c]{
    \begin{tabular}{ccccccc}
    \hline
         & & FPR at & Detection &  & AUPR & AUPR\\
        ID & OOD & 95\% TPR $\downarrow$ & Error $\downarrow$ & AUROC $\uparrow$ & Out $\uparrow$ & In $\uparrow$\\
        \cline{3-7}
         & & \multicolumn{5}{c}{ODIN/DAGMM/DOODLER}\\
        \hline
        \hline
MNIST&F-MNIST&0.49/0.35/0.19&0.27/0.2/0.12&0.89/0.92/0.96&0.89/0.92/0.96&0.88/0.92/0.96\\
        \hline
    \end{tabular}}
    \caption{Example Experimental Result}
    \hrulefill
    \label{tab:res_simp_ex}
\end{table}

Following other papers on the subject \cite{DBLP:journals/corr/abs-1910-10307, DBLP:journals/corr/LiangLS17}, our experiments were done by selecting two available Computer Vision datasets, and declaring one to be the In-Distribution, and the other Out-Of-Distribution. Then a VAE was trained on the training portion of the ID dataset, and used to classify samples from the testing portions of both the ID and OOD datasets. For example, Table \ref{tab:res_simp_ex} shows an experiment where a VAE was trained on the training portion of the MNIST dataset, and then was tasked with classifying testing MNIST samples as In-Distribution, versus Fashion-MNIST samples as Out-Of-Distribution. The metrics shown will be explained in the Results and Comparisons section.\\
\\
The datasets we selected include MNIST \cite{lecun-mnisthandwrittendigit-2010}, Fashion-MNIST \cite{DBLP:journals/corr/abs-1708-07747}, Omniglot \cite{Lake1332}, CIFAR-100 \cite{cifar100cite}, TinyImagenet \cite{deng2009imagenet, Le2015TinyIV}, SVHN \cite{netzer2011reading}, and CelebA \cite{liu2015faceattributes}. These datasets were collected together into ``simple" and ``complex" groups, with MNIST, F-MNIST, and Omniglot in the former, and CIFAR-100, TinyImagenet, SVHN, and CelebA in the latter. Each dataset had its own VAE trained for it, which was then tested with DOODLER for its ability to detect samples from every other dataset in the group as Out-Of-Distribution. Additionally, in both groups, random Gaussian noise clipped to $[0, 1]$ and Uniform noise from $[0, 1]$ were added as purely OOD samples.\\
\\
Implementations of ODIN and DAGMM were used as comparisons, with models similar in capability and with equivalent training regimens to the VAEs used for DOODLER.


\section{Results and Comparisons}

\begin{table}[t]
    \centering
    \makebox[\textwidth][c]{
    \begin{tabular}{ccccccc}
    \hline
         & & FPR at & Detection &  & AUPR & AUPR\\
        ID & OOD & 95\% TPR $\downarrow$ & Error $\downarrow$ & AUROC $\uparrow$ & Out $\uparrow$ & In $\uparrow$\\
        \cline{3-7}
         & & \multicolumn{5}{c}{ODIN/DAGMM/DOODLER}\\
        \hline
        \hline
MNIST&F-MNIST&0.49/0.35/\textbf{0.19}&0.27/0.2/\textbf{0.12}&0.89/0.92/\textbf{0.96}&0.89/0.92/\textbf{0.96}&0.88/0.92/\textbf{0.96}\\
&Omniglot&0.33/0.21/\textbf{0.07}&0.19/0.13/\textbf{0.06}&0.93/0.96/\textbf{0.98}&0.93/0.97/\textbf{0.98}&0.93/0.96/\textbf{0.98}\\
&Gaussian&0.12/0.17/\textbf{0.01}&0.09/0.11/\textbf{0.03}&0.98/0.97/\textbf{1.0}&0.98/0.97/\textbf{1.0}&0.98/0.97/\textbf{1.0}\\
&Uniform&0.12/0.17/\textbf{0.02}&0.08/0.11/\textbf{0.03}&0.98/0.96/\textbf{1.0}&0.98/0.96/\textbf{1.0}&0.98/0.96/\textbf{0.99}\\
        \hline
F-MNIST&MNIST&0.49/0.23/\textbf{0.01}&0.27/0.14/\textbf{0.03}&0.88/0.95/\textbf{1.0}&0.88/0.95/\textbf{1.0}&0.88/0.95/\textbf{1.0}\\
&Omniglot&0.59/0.18/\textbf{0.0}&0.32/0.12/\textbf{0.03}&0.85/0.97/\textbf{1.0}&0.86/0.97/\textbf{1.0}&0.85/0.96/\textbf{1.0}\\
&Gaussian&\textbf{0.02}/\textbf{0.02}/\textbf{0.02}&\textbf{0.03}/0.04/\textbf{0.03}&0.99/\textbf{1.0}/0.99&\textbf{0.99}/\textbf{0.99}/\textbf{0.99}&\textbf{0.99}/\textbf{0.99}/\textbf{0.99}\\
&Uniform&0.02/\textbf{0.01}/0.03&0.04/\textbf{0.03}/0.04&0.99/\textbf{1.0}/0.99&0.99/\textbf{1.0}/0.99&0.99/\textbf{1.0}/0.99\\
        \hline
Omniglot&MNIST&0.22/0.43/\textbf{0.04}&0.14/0.24/\textbf{0.04}&0.96/0.89/\textbf{0.99}&0.95/0.9/\textbf{0.99}&0.96/0.89/\textbf{0.99}\\
&F-MNIST&0.26/0.56/\textbf{0.05}&0.16/0.3/\textbf{0.05}&0.94/0.87/\textbf{0.99}&0.95/0.88/\textbf{0.99}&0.94/0.86/\textbf{0.99}\\
&Gaussian&0.03/0.08/\textbf{0.0}&0.04/0.07/\textbf{0.03}&0.99/0.98/\textbf{1.0}&0.99/0.98/\textbf{1.0}&0.99/0.98/\textbf{1.0}\\
&Uniform&0.03/0.06/\textbf{0.0}&0.04/0.05/\textbf{0.03}&0.99/0.99/\textbf{1.0}&0.99/0.99/\textbf{1.0}&0.99/0.99/\textbf{1.0}\\
        \hline
    \end{tabular}}
    \caption{Result comparisons on visually simple datasets}
    \label{tab:res_simp}
    \hrulefill
\end{table}

\begin{table}[t]
    \centering
    \makebox[\textwidth][c]{
    \begin{tabular}{ccccccc}
    \hline
         & & FPR at & Detection &  & AUPR & AUPR\\
        ID & OOD & 95\% TPR $\downarrow$ & Error $\downarrow$ & AUROC $\uparrow$ & Out $\uparrow$ & In $\uparrow$\\
        \cline{3-7}
         & & \multicolumn{5}{c}{ODIN/DAGMM/DOODLER}\\
        \hline
        \hline
CIFAR-100&TinyImagenet&0.81/0.63/\textbf{0.5}&0.43/0.34/\textbf{0.28}&0.73/0.81/\textbf{0.86}&0.73/0.79/\textbf{0.86}&0.71/0.81/\textbf{0.86}\\
&SVHN&0.84/0.93/\textbf{0.8}&0.44/0.49/\textbf{0.42}&0.68/0.57/\textbf{0.71}&0.66/0.55/\textbf{0.69}&0.67/0.55/\textbf{0.7}\\
&CelebA&0.84/0.82/\textbf{0.47}&0.45/0.44/\textbf{0.26}&0.67/0.72/\textbf{0.88}&0.66/0.7/\textbf{0.88}&0.66/0.71/\textbf{0.88}\\
&Gaussian&0.03/0.02/\textbf{0.01}&0.04/0.04/\textbf{0.03}&0.99/\textbf{1.0}/\textbf{1.0}&0.99/0.99/\textbf{1.0}&0.99/0.99/\textbf{1.0}\\
&Uniform&0.02/0.02/\textbf{0.0}&0.04/0.04/\textbf{0.03}&0.99/\textbf{1.0}/\textbf{1.0}&0.99/0.99/\textbf{1.0}&0.99/0.99/\textbf{1.0}\\
        \hline
TinyImagenet&CIFAR-100&0.89/0.75/\textbf{0.64}&0.47/0.4/\textbf{0.34}&0.62/0.73/\textbf{0.82}&0.61/0.71/\textbf{0.82}&0.61/0.72/\textbf{0.82}\\
&SVHN&0.78/0.79/\textbf{0.67}&0.41/0.42/\textbf{0.36}&0.74/0.74/\textbf{0.78}&0.73/0.74/\textbf{0.78}&0.72/0.73/\textbf{0.77}\\
&CelebA&0.61/\textbf{0.53}/0.78&0.33/\textbf{0.29}/0.41&0.81/\textbf{0.86}/0.72&0.81/\textbf{0.86}/0.71&0.81/\textbf{0.86}/0.72\\
&Gaussian&0.03/0.04/\textbf{0.01}&0.04/0.05/\textbf{0.03}&0.99/0.99/\textbf{1.0}&0.99/0.99/\textbf{1.0}&0.99/0.99/\textbf{1.0}\\
&Uniform&0.03/0.03/\textbf{0.01}&0.04/0.04/\textbf{0.03}&0.99/0.99/\textbf{1.0}&0.99/0.99/\textbf{1.0}&0.99/0.99/\textbf{1.0}\\
        \hline
SVHN&CIFAR-100&0.58/0.45/\textbf{0.32}&0.32/0.25/\textbf{0.19}&0.85/0.9/\textbf{0.93}&0.86/0.9/\textbf{0.92}&0.84/0.9/\textbf{0.93}\\
&TinyImagenet&0.23/0.41/\textbf{0.14}&0.14/0.23/\textbf{0.1}&0.95/0.91/\textbf{0.97}&0.95/0.91/\textbf{0.97}&0.95/0.91/\textbf{0.97}\\
&CelebA&0.35/0.23/\textbf{0.06}&0.2/0.14/\textbf{0.05}&0.91/0.96/\textbf{0.99}&0.91/0.96/\textbf{0.99}&0.91/0.96/\textbf{0.99}\\
&Gaussian&\textbf{0.01}/\textbf{0.01}/\textbf{0.01}&\textbf{0.03}/\textbf{0.03}/\textbf{0.03}&\textbf{1.0}/\textbf{1.0}/\textbf{1.0}&0.99/0.99/\textbf{1.0}&0.99/0.99/\textbf{1.0}\\
&Uniform&0.02/0.02/\textbf{0.01}&\textbf{0.03}/0.04/\textbf{0.03}&\textbf{1.0}/0.99/\textbf{1.0}&\textbf{1.0}/0.99/\textbf{1.0}&\textbf{1.0}/0.99/\textbf{1.0}\\
        \hline
CelebA&CIFAR-100&0.86/\textbf{0.78}/0.9&0.46/\textbf{0.41}/0.47&0.67/\textbf{0.73}/0.6&0.67/\textbf{0.72}/0.6&0.65/\textbf{0.72}/0.59\\
&TinyImagenet&0.64/\textbf{0.54}/0.73&0.35/\textbf{0.29}/0.39&0.81/\textbf{0.85}/0.75&0.8/\textbf{0.84}/0.75&0.8/\textbf{0.85}/0.75\\
&SVHN&\textbf{0.61}/0.82/0.89&\textbf{0.33}/0.44/0.47&\textbf{0.81}/0.7/0.61&\textbf{0.8}/0.69/0.6&\textbf{0.81}/0.68/0.61\\
&Gaussian&0.02/0.02/\textbf{0.01}&\textbf{0.03}/0.04/\textbf{0.03}&\textbf{1.0}/\textbf{1.0}/\textbf{1.0}&\textbf{1.0}/0.99/\textbf{1.0}&\textbf{1.0}/0.99/\textbf{1.0}\\
&Uniform&0.02/0.02/\textbf{0.01}&\textbf{0.03}/\textbf{0.03}/\textbf{0.03}&\textbf{1.0}/\textbf{1.0}/\textbf{1.0}&\textbf{1.0}/0.99/\textbf{1.0}&\textbf{1.0}/0.99/\textbf{1.0}\\
        \hline
    \end{tabular}}
    \caption{Results comparisons on more visually complex datasets}
    \label{tab:res_comp}
    \hrulefill
\end{table}

When doing analyses of the results, we assumed the prior probabilities $p(x_i \sim \mathscr{D}_{ID}),\ p(x_i \not \sim \mathscr{D}_{ID})$ of receiving an In-Distribution and an Out-Of-Distribution sample, respectively, to both be $\frac{1}{2}$.\\
\\
The metrics used are taken from \cite{DBLP:journals/corr/HendrycksG16c, DBLP:journals/corr/LiangLS17, DBLP:journals/corr/abs-1910-10307}. They include:

\begin{itemize}
    \item The FPR at 95\% TPR, which is the probability of an ID sample being accidentally declared OOD when a real OOD sample has a 95\% chance of being detected. This should be minimized.
    \item Detection Error is the over-all probability of any given sample being misclassified when at a 95\% True Positive Rate. This should be minimized.
    \item Area Under the Receiver Operating Characteristic curve is the over-all probability that an OOD sample receives a higher metric than an ID sample, in the case of DOODLER a higher reconstruction error. This should be maximized.
    \item Area Under the Precision-Recall curve represents balancing a low rate of false positives with a high rate of collecting true positives. Both the AUPR calculated with OOD samples being positive and ID samples being positive are provided. Both should be maximized.
\end{itemize}

\begin{figure}[h]
    \centering
    \includegraphics[width=0.6\textwidth]{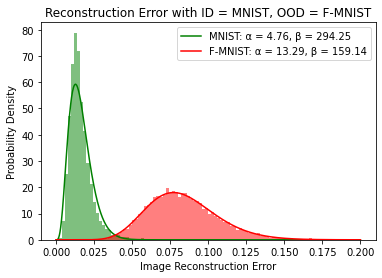}
    \caption{Reconstruction errors compared between the MNIST and Fashion-MNIST datasets with a VAE trained on MNIST}
    \label{fig:mnistvsfmnist}
    \hrulefill
\end{figure}

\begin{figure}[h]
    \centering
    \includegraphics[width=0.6\textwidth]{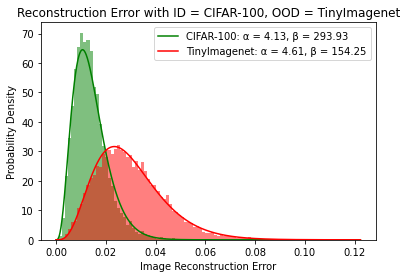}
    \caption{Reconstruction errors compared between the CIFAR-100 and TinyImagenet datasets with a VAE trained on CIFAR-100}
    \label{fig:cifar100vstinyimagenet}
    \hrulefill
\end{figure}

\begin{figure}
    \centering
    \includegraphics[width=0.6\textwidth]{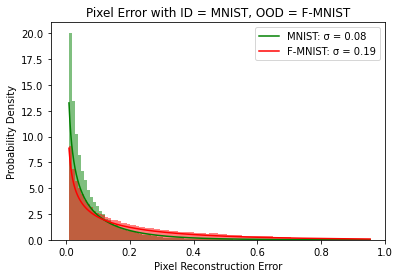}
    \caption{Pixel-wise errors compared between the MNIST and Fashion-MNIST datasets with a VAE trained on MNIST}
    \label{fig:mnistvsfmnistpixels}
    \hrulefill
\end{figure}

\begin{figure}
    \centering
    \includegraphics[width=0.6\textwidth]{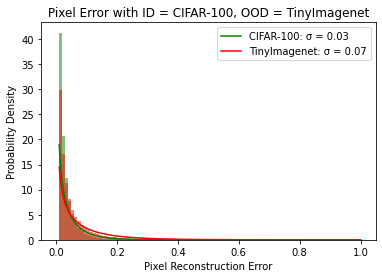}
    \caption{Pixel-wise errors compared between the CIFAR-100 and TinyImagenet datasets with a VAE trained on CIFAR-100}
    \label{fig:cifar100vstinyimagenetpixels}
    \hrulefill
\end{figure}

\begin{figure}
    \centering
    \includegraphics[width=0.6\textwidth]{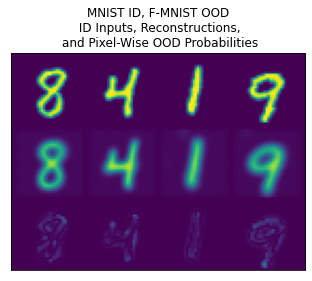}
    \includegraphics[width=0.6\textwidth]{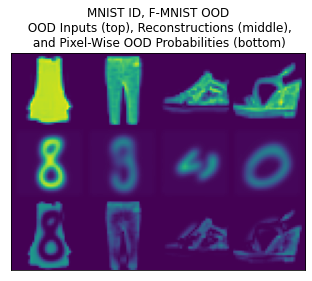}
    \caption{Comparison of reconstructions and pixel-wise OOD probabilities between MNIST and Fashion-MNIST samples using a VAE trained on MNIST}
    \label{fig:mnistvsfmnistsamples}
    \hrulefill
\end{figure}

\begin{figure}
    \centering
    \includegraphics[width=0.6\textwidth]{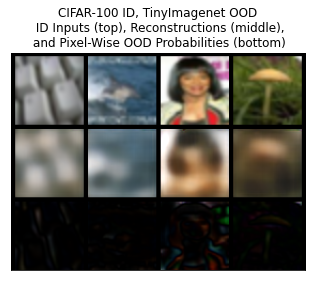}
    \includegraphics[width=0.6\textwidth]{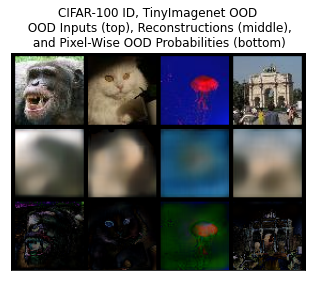}
    \caption{Comparison of reconstructions and pixel-wise OOD probabilities between CIFAR-100 and TinyImagenet samples using a VAE trained on CIFAR-100}
    \label{fig:cifar100vstinyimagenetsamples}
    \hrulefill
\end{figure}

\noindent Looking at Tables \ref{tab:res_simp} and \ref{tab:res_comp}, we can see that DOODLER competes very well against ODIN and DAGMM, often superseding them. An example of this high performance is seen in the graph of reconstruction errors on MNIST and Fashion-MNIST with a VAE trained for MNIST, given in Figure \ref{fig:mnistvsfmnist}, which highlights the differences between the reconstruction error distributions. In that figure, we see very little overlap, meaning that determining which inputs belong to which dataset based on the reconstruction error yields good results. Similarly, Figure \ref{fig:cifar100vstinyimagenet} shows the differences in the reconstruction errors between CIFAR-100 and TinyImagenet with a VAE trained on CIFAR-100. While the differences are less stark, they are still very significant, allowing DOODLER to reach an AUROC of 0.86.\\
\\
Looking at Figures \ref{fig:mnistvsfmnistpixels} and \ref{fig:cifar100vstinyimagenetpixels}, which contain the $\chi^2$ distributions regressed from the actual error contributions of individual pixels, shows us that the actual difference between how accurate the VAE is on any given pixel of an In-Distribution versus an Out-Of-Distribution input is relatively subtle. Rather, the large differences in over-all image reconstruction error are due to the small per-pixel differences adding up over a very large number of pixels, producing very different per-image distributions.\\
\\
Additionally, Figures \ref{fig:mnistvsfmnistsamples} and \ref{fig:cifar100vstinyimagenetsamples} provide examples of inputs, reconstructions, and the estimated per-pixel Out-Of-Distribution likelihoods from both In-Distribution and OOD samples. Even though the reconstructions on the visually complex datasets are quite cloudy, they do align relatively well with the inputs in shape and color on the ID samples, while failing on the OOD samples, producing significantly greater error. Looking at the pixel-wise OOD likelihood estimation, it becomes clear to see that the bright spots align themselves closely with features in the image, representing the objects that the VAE was unfamiliar with from training.\\
\\
This means that the intuitions developed earlier about the functionality of DOODLER being based on a failure to reconstruct unseen data bear out experimentally, because on a fundamental level, what is happening is that the VAEs trained on the In-Distribution are failing to reconstruct objects they haven't seen before in the Out-Of-Distribution inputs. In a very real sense, this means that DOODLER is performing Out-Of-Distribution Detection to the letter, because it highlights objects that don't exist within the In-Distribution.

\section{Conclusions and Future Work}

While DOODLER represents a step forward for Out-Of-Distribution detection, work remains to be done. This includes placing theoretical bounds on its error rate, understanding the reconstruction error distributions more wholly than is possible with Gamma distributions, and increasing the saliency of pixel-wise segmentations. Additionally, it might be worthwhile to try and study the connection between how the Variational Auto-Encoder builds its latent space, and the resulting reconstruction errors, to help establish what exactly in the sample has resulted in detection as an OOD input. It might also be worthwhile to pair the VAE with some kind of Generative Adversarial framework \cite{goodfellow2014generative}, to try and improve the sharpness of the reconstructions. It may also be possible to further improve on OOD Detection by combining elements from multiple methodologies, including DOODLER, or by using a an ensemble of them to cover each other's weaknesses.\\
\\
There is also work to be done to study how DOODLER behaves when provided with adversarial inputs. For example, can an adversarial attack actually decrease the reconstruction error of an input? This could be a potential avenue of attack, if an adversary wanted to provide an anomalous input without a human operator being notified.\\
\\
But, to conclude, DOODLER represents a novel framework for Out-Of-Distribution Detection, introducing a functional perspective to the problem, improving on aspects of earlier methods, and providing a basis on which to pursue functionality like image segmentation. This constitutes a step forwards in the pursuit of making Machine Learning robust and trustworthy for real-world applications.

\bibliographystyle{plain}
\bibliography{references}

\newpage

\section{Example Figures}

\end{document}